\documentclass{article}
\usepackage{amsmath,graphicx,url}
\usepackage{amsfonts}
\usepackage{amsmath}
\usepackage{amsthm}

\newtheorem{proposition}{Proposition}
\newtheorem{lemma}{Lemma}
\newtheorem{corollary}{Corollary}
\usepackage{amssymb}
\usepackage{pifont}

\usepackage{makecell}

\usepackage{booktabs}

\usepackage{enumitem}
\usepackage[margin=1in]{geometry}
\setlist{nosep, leftmargin=14pt}
\usepackage{authblk}

\usepackage{mwe} 

\title{Bias-Aware Conformal Prediction for Metric-Based Imaging Pipelines}

\author{Matt Y. Cheung$^1$}
\author{Tucker J. Netherton$^2$}
\author{Laurence E. Court$^2$}
\author{Ashok Veeraraghavan$^1$}
\author{Guha Balakrishnan$^1$}
\affil{$^1$ Department of Electrical $\&$ Computer Engineering, Rice University, Houston TX}
\affil{$^2$ Department of Radiation Physics, The University of Texas MD Anderson Cancer Center, Houston TX}

\begin{document}
\maketitle

\begin{abstract}
Reliable confidence measures of metrics derived from medical imaging reconstruction pipelines would improve the standard of decision-making in many clinical workflows. Conformal Prediction (CP) provides a robust framework for producing calibrated prediction intervals, but standard CP formulations face a critical challenge in the imaging pipeline: common mismatches between image reconstruction objectives and downstream metrics can introduce systematic prediction deviations from ground truth values, known as bias. These biases in turn compromise the efficiency of prediction intervals, which is a problem that has been unexplored in the CP literature. In this study, we formalize the behavior of symmetric (where bounds expand equally in both directions) and asymmetric (where bounds expand unequally) formulations for common non-conformity scores in CP in the presence of bias, and argue that this measurable bias must inform the choice of CP formulation. We theoretically and empirically demonstrate that symmetric intervals are inflated by a factor of two times the magnitude of bias while asymmetric intervals remain unaffected by bias, and provide conditions under which each formulation produces tighter intervals. 
We empirically validated our theoretical analyses on sparse-view CT reconstruction for downstream radiotherapy planning. Our work enables users of medical imaging pipelines to proactively select optimal CP formulations, thereby improving interval length efficiency for critical downstream metrics.
\end{abstract}
%
%

\begin{figure}[t]
    \centering
    \includegraphics[scale=0.1]{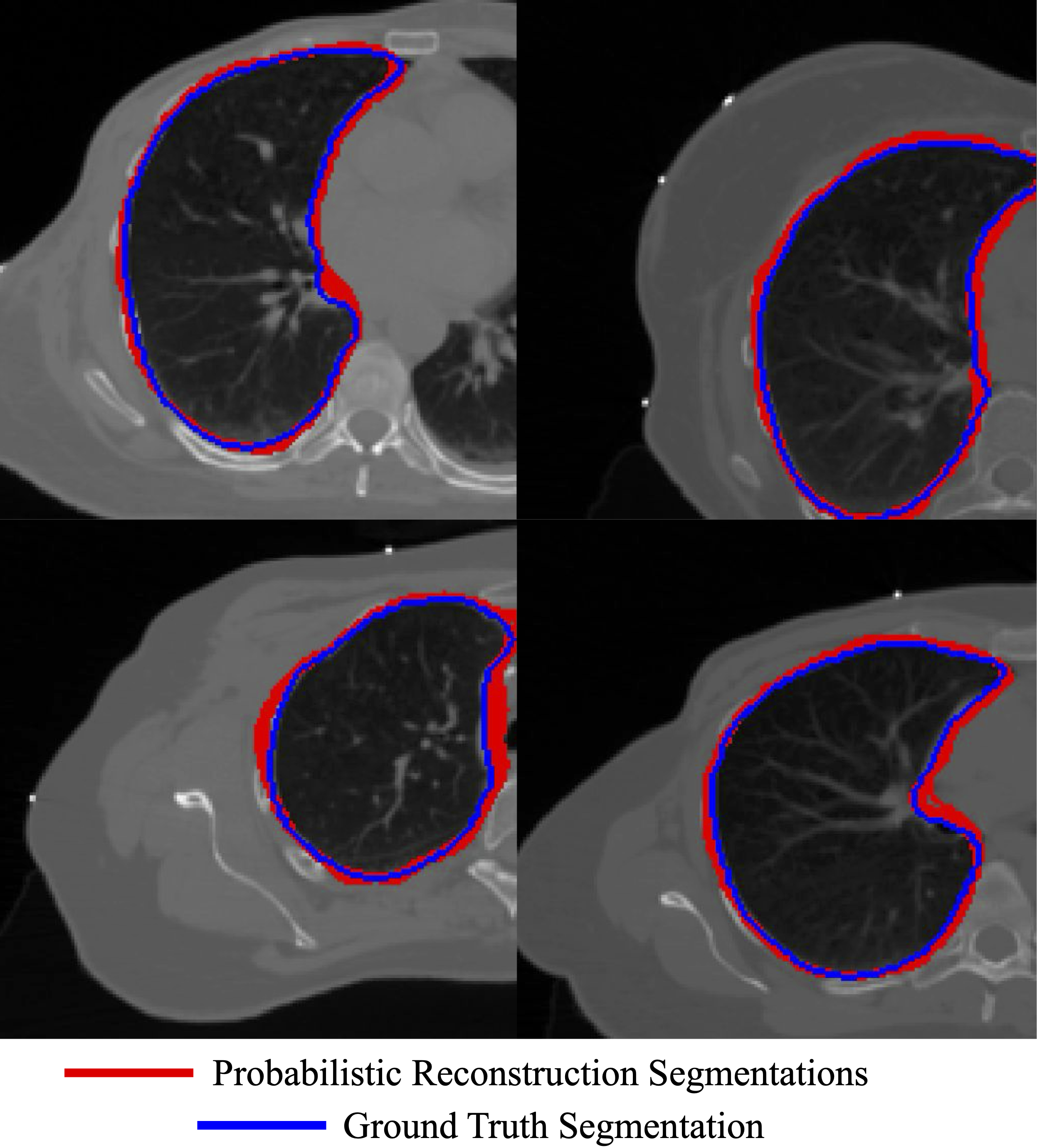}
    \caption{\textbf{Biases in sparse-CT reconstructions generated from NAF for a downstream lung segmentation task.} We show slices of CT volumes from four different patients. Each slice is overlaid with contours of the right lung generated from 10 probabilistic CT reconstructions (red) and ground truth contours (blue). The reconstructed right lung segmentations consistently overestimate organ volumes compared to the ground truth segmentations.}
    \label{fig:seg}
\end{figure}

\section{Introduction}

In many medical imaging pipelines, high-stakes clinical decisions are based on quantitative downstream metrics, such as tumor volume, derived from reconstructed images. 
Therefore, deploying the pipeline safely requires not just a point prediction, but a reliable measure of confidence. 
Conformal Prediction (CP) offers a robust, distribution-free framework for producing calibrated prediction intervals~\cite{angelopoulos2021gentle,fontana2023conformal,papadopoulos2002inductive}.
Split CP, for example, uses a held-out \emph{calibration} dataset to compute \emph{non-conformity scores} that quantify how ``unusual'' new predictions are, then adjusts the prediction interval via an empirical quantile of these scores.
Scores can take symmetric (where bounds are adjusted equally in both direction) or asymmetric (where bounds are adjusted unequally)~\cite{linusson2014signed,romano2019conformalized}.
In practice, we desire the tightest possible intervals that maintain valid coverage.
However, interval tightness can be severely compromised by the medical imaging pipeline itself.
Because the reconstruction algorithm is optimized for pixel-wise losses rather than the downstream metric, mismatched objectives between pipeline components can introduce systematic biases or artifacts, leading the model’s predictions to  \emph{systematically deviate} from ground truth values.
This bias can be defined as the mean expected difference between predictions $\hat{Y}_i$ and ground truth metrics $Y_i$: $b = \frac{1}{n} \sum_{i=1}^n (\mathbb{E}[\hat Y_i] - Y_i)$.
While the effect of bias has been well studied in classical statistics, it is underexplored in CP and  medical imaging.

In this paper, we argue that bias should inform the choice of CP formulation.
While existing theory suggests asymmetric intervals are wider compared to symmetric intervals~\cite{romano2019conformalized}, we show this is not necessarily the case in the presence of bias.
We expand the theoretical understanding of CP for absolute residual~\cite{lei2018distribution} and quantile-adjusted~\cite{romano2019conformalized} non-conformity scores by analyzing their behavior under prediction bias.
Specifically, we formalize:
\begin{enumerate}
    \item The upper bound of symmetric interval lengths increases by $2|b|$ (Prop.~\ref{prop:symL}),
    \item Asymmetric interval lengths are not affected by bias (Prop.~\ref{prop:asymIndepBias}), and
    \item The condition when asymmetric formulations produce more efficient interval lengths than  symmetric formulations (Cor.~\ref{cor:SymVSAsymLengths}).
\end{enumerate}
We validate our theory on sparse-view CT reconstruction tasks.
Our results are significant for two reasons.
First, they provide a formal explanation for empirical observations that asymmetric intervals can be tighter than symmetric ones~\cite{linusson2014signed,wang2023conformal,cheung2024metric}---a phenomenon we attribute to unaddressed bias.
Second, our results empower practitioners to account for bias and \emph{proactively select} which formulation to improve interval efficiency for downstream metrics.

\section{Background: Split CP}
We focus on a ``split'' CP setup~\cite{papadopoulos2002inductive,lei2018distribution} in this work, but the same theoretical analysis can be applied to other CP forms and extensions~\cite{fontana2023conformal,barber2021predictive}.
In split CP, we assume a calibration dataset $D_C=\{(\hat Y_1, Y_1),...,(\hat Y_n, Y_n)\}$ and test point $\hat Y_{n+1}$, where $\hat Y_i$ and $Y_i$ represent the $i$-th prediction and ground truth values. The calibration data is separate from a training dataset used to train the ML algorithm of interest. The calibration dataset and test point are assumed to be exchangeable. The goal of CP is to construct a prediction interval $C(\hat Y_{n+1})=[L(\hat Y_{n+1}), U(\hat Y_{n+1})]$ for $\hat Y_{n+1}$, where $L(\hat Y_{n+1}),U(\hat Y_{n+1})\in \mathbb{R}$ are lower and upper bounds, such that $\mathbb{P}[Y_{n+1} \in C(\hat Y_{n+1})]\geq 1-\alpha$, for some user-specified mis-coverage rate $\alpha\in (0,1)$. 
To compute symmetric intervals, we perform the following steps. 
First, for each data point in the calibration set $D_C$, we compute non-conformity scores $S=\{s_1,...,s_n\}$. 
Next, we compute the $(1-\alpha)$-th empirical quantile of the non-conformity scores $q=Q_{1-\hat \alpha}(S)$, where $\hat \alpha=\frac{\lfloor \alpha (n+1)\rfloor}{n}$ denotes the finite-sample adjusted mis-coverage rate.
Finally, we adjust the predictions of the test data using $q$ to achieve valid prediction sets. 
This algorithm provides marginal coverage: on average, the prediction sets contain ground truth $(1-\alpha)\%$ of the time. More rigorously, based on key CP results:
\begin{lemma}
    \label{prop:conformal}
    Let $(\hat Y_i, Y_i) \in \mathbb{R} \times \mathbb{R}, i=1,...,n+1$ be exchangeable random variables. 
    Assume that a predictor $f$ has been trained on a proper training set independent of and exchangeable with these $n+1$ points.
    Consider a calibration set ${(\hat Y_i, Y_i)}_{i=1}^n$ and a fresh test point $\hat Y_{n+1}$.
    Let $s_i$ be a non-conformity score computed using the predictor $f$ for $i=1,...,n+1$.
    Let $q=Q_{1-\hat \alpha}(\{s_i\}_{i=1}^n)$ be the $\lceil(1-\alpha)(n+1)\rceil$-th smallest value of $\{s_i\}_{i=1}^n$ and $C(\hat Y_{n+1})=\{y \in \mathbb{R}: s_{n+1} \leq q\}$ be the prediction set for the test point $\hat Y_{n+1}$.
    Then, for any $\alpha \in (0,1)$:
    \begin{enumerate}
        \item $\mathbb{P}[Y_{n+1} \in C(\hat Y_{n+1})] \geq 1-\alpha$ and
        \item $\mathbb{P}[Y_{n+1} \in C(\hat{Y}_{n+1})] \leq 1-\alpha + \frac{1}{n+1}$ if random variables $Y_1,...,Y_{n+1}$ are almost surely distinct
    \end{enumerate}
    
    \noindent Proof: This is a standard result of split CP. See \cite{lei2018distribution,vovk2005algorithmic,tibshirani2019conformal,oliveira2024split}
\end{lemma}
Many non-conformity scores exist~\cite{kato2023review}, including absolute residuals ($L_1$)~\cite{papadopoulos2002inductive} and quantile-based (Conformalized Quantile Regression or CQR ~\cite{romano2019conformalized}) scores. Such scores can be extended to asymmetric formulations~\cite{linusson2014signed,romano2019conformalized} by computing $\alpha_{lo}$-th and $(1-\alpha_{hi})$-th empirical quantiles of the conformity scores, where $\alpha_{lo}$ and $\alpha_{hi}$ are lower and upper mis-coverage rates and the finite sample adjustment for $1-\alpha$ is $\lceil(1-\alpha)(n+1)\rceil/n$. 
It is easy to see that when $\alpha_{lo}+\alpha_{hi}=\alpha$, the asymmetric case yields empirically larger coverage based on the finite sample adjustment.
The reason is due to the ``rounding effect'' of the ceiling function, which tends to push the empirical quantiles toward more extreme values, especially when $n$ is small. Since the asymmetric case deals with two separate quantiles, this effect is compounded, leading to a prediction set that empirically offers larger coverage. However, as we will see, symmetric interval lengths are not always shorter than asymmetric ones in the presence of bias. 

\section{Theoretical Analysis}
\label{sec:theory}

We assume biased predictions of the form $\hat Y_i^b = \hat Y_i^0 + b$
where $b\in\mathbb{R}$ is a scalar additive bias applied uniformly across the predictions $\{\hat Y_i^0\}_{i=1}^n$ relative to the ground-truth values $\{Y_i\}_{i=1}^n$. We consider symmetric nonconformity scores of the canonical form: $s_i^b \;=\; \max\big( f_{lo}(\hat Y_i^b) - Y_i,\; Y_i - f_{hi}(\hat Y_i^b)\big)$
where $f_{lo}$ and $f_{hi}$ denote the lower and upper adjustment functions that map a (possibly set-valued) prediction $\hat Y_i^b$ to scalar lower/upper reference values used in the score. 
Two important cases are: $L_1$ (absolute residual) scores and CQR (quantile-based) scores.
For $L_1$, the prediction is simply a point estimate $\hat Y_i^b\in\mathbb{R}$, adjustment functions are $f_{lo}(\hat Y_i^b)=f_{hi}(\hat Y_i^b)=\hat Y_i^b$ and score is $s_i^b = |Y_i-\hat Y_i^b| = \max(\hat Y_i^b - Y_i,\; Y_i - \hat Y_i^b)$.
For CQR, the adjustments $f_{lo}$ and $f_{hi}$ either produced by the model or computed by applying an empirical quantile operator $Q_\alpha(\cdot)$ to this sample set: $f_{lo}(\hat Y_i^b)=Q_{\alpha_{lo}}(\hat Y_i^b)$ and $ f_{hi}(\hat Y_i^b)=Q_{1-\alpha_{hi}}(\hat Y_i^b)$. The CQR nonconformity score then becomes $s_i^b=\max( Q_{\alpha_{lo}}(\hat Y_i^b) - Y_i,\; Y_i - Q_{1-\alpha_{hi}}(\hat Y_i^b))$.

A key property we exploit is the \emph{translation equivariance} of the empirical quantile operator: for any scalar $b$: $Q_\alpha(\hat{Y_i}+b) = Q_\alpha(\hat{Y}_i)+b$.
Consequently, when the bias is additive, the adjustments shift by $b$: $f_{lo}(\hat Y_i^b)=f_{lo}(\hat Y_i^0)+b$ and $f_{hi}(\hat Y_i^b)=f_{hi}(\hat Y_i^0)+b$,
provided $f_{lo}$ and $f_{hi}$ are obtained via translation-equivariant operators (e.g., identity for $L_1$, empirical quantiles for CQR).
The conformal adjustment is obtained as the empirical $(1-\alpha)$ quantile of the scores: $q^b=Q_{1-\hat\alpha}(\{s_i^b\}_{i=1}^n)$. The resulting symmetric prediction interval for a new biased prediction $\hat Y_{n+1}^b$ is $C(\hat{Y}_{n+1}^b)=[f_{lo}(\hat{Y}_{n+1}^b) - q^b,f_{hi}(\hat{Y}_{n+1}^b)+q^b]$
and its length is $L_{sym}(\hat Y_{n+1}^b)=f_{hi}(\hat Y_{n+1}^b) - f_{lo}(\hat Y_{n+1}^b) + 2q^b$. Thus, upper bound for symmetric interval lengths under bias is given by the following proposition:
\begin{proposition}
    \label{prop:symL}
    Given biased predictions for a fresh test point $\hat Y_{n+1}^b = \hat Y_{n+1}^0+b$, the upper bound on prediction interval lengths of non-conformity scores described by the canonical form is: $L_{sym}(\hat Y_{n+1}^b) \leq L_{sym}(\hat Y_{n+1}^0)+2|b|$, where $L_{sym}(\hat Y_{n+1}^b)$ and $L_{sym}(\hat Y_{n+1}^0)$ are the interval lengths computed using symmetric adjustments for predictions with and without bias.

    \noindent \textbf{Proof Sketch.} By translation equivariance, the upper and lower adjustments shift by the bias \(b\), so the interval span remains unchanged, \(f_{hi}(\hat Y_{n+1}^b)-f_{lo}(\hat Y_{n+1}^b)=f_{hi}(\hat Y_{n+1}^0)-f_{lo}(\hat Y_{n+1}^0)\), and since each calibration score satisfies \(|s_i^b - s_i^0|\le |b|\), the empirical quantile satisfies \(q^b \le q^0 + |b|\). Combining these gives $L_{sym}(\hat Y_{n+1}^b) = f_{hi}(\hat Y_{n+1}^b)-f_{lo}(\hat Y_{n+1}^b) + 2q^b \le f_{hi}(\hat Y_{n+1}^0)-f_{lo}(\hat Y_{n+1}^0) + 2(q^0 + |b|)=L_{sym}(\hat Y_{n+1}^0) + 2|b|$ as claimed.
\end{proposition}

We find the upper bounds of symmetric interval lengths increase \emph{linearly} with the magnitude of bias.

Next, we extend the discussion to asymmetric adjustments. Let $(s_{i,lo}^b, s_{i,hi}^b) \;=\; \big(f_{lo}(\hat Y_i^b)-Y_i,\; Y_i - f_{hi}(\hat Y_i^b)\big)$, where $s_{i,lo}^b$ and $s_{i,hi}^b$, denote the lower and upper non-conformity scores for biased predictions $\hat{Y}_i^b=\hat{Y}_i^0+b$.
As in the symmetric case, the empirical lower and upper adjustments are computed via the corresponding quantiles of the calibration scores:
$q_{lo}^b=Q_{1-\hat{\alpha}_{lo}}(\{s_{i,lo}^b\}_{i=1}^n)$ and $q_{hi}^b=Q_{1-\hat{\alpha}_{hi}}(\{s_{i,hi}^b\}_{i=1}^n)$.  By translation equivariance of $f_{lo}$ and $f_{hi}$ and the quantile operator, these adjustments satisfy $q_{lo}^b = q_{lo}^0 + |b|$ and $q_{hi}^b = q_{hi}^0 + |b|$.  The resulting asymmetric prediction interval for a new biased prediction $\hat Y_{n+1}^b$ is $C(\hat{Y}_{n+1}^b)=[f_{lo}(\hat{Y}_{n+1}^b)-q_{lo}^b,f_{hi}(\hat{Y}_{n+1}^b) + q_{hi}^b]$

\begin{table*}[t]
\centering
\begin{tabular}{@{}cccccccc@{}}
\toprule
Metric & $L_{sym}^b (\downarrow)$ & $L_{asym}^b(\downarrow)$ & $L_{sym}^0$ & $|b|$ & $L_{asym}^0-L_{sym}^0$ & $C_{sym}$ & $C_{asym}$ \\ \midrule
Heart $D_{35}$ (Gy) & 2.07e-01 & \textbf{1.89e-01} & 1.89e-01 & 1.16e-02 & -1.83e-03 & 0.901 & 0.902 \\
Heart Volume (cm$^3$) & \textbf{3.03e+02} & 4.50e+02 & 3.01e+02 & 1.63e+00 & 1.51e+02 & 0.905 & 0.904 \\
Right Lung $V_{20}$ (cm$^3$) & 6.84e-02 & \textbf{6.48e-02} & 6.41e-02 & 8.21e-03 & 1.10e-03 & 0.901 & 0.898 \\
Right Lung $D_{35}$ (Gy) & 1.02e+01 & \textbf{9.29e+00} & 9.71e+00 & 7.58e-01 & -5.24e-01 & 0.902 & 0.899 \\
Right Lung Volume (cm$^3$) & 3.04e+02 & \textbf{2.00e+02} & 1.64e+02 & 8.66e+01 & 3.59e+01 & 0.925 & 0.904 \\
Left Lung Volume (cm$^3$) & 3.05e+02 & \textbf{2.45e+02} & 2.09e+02 & 6.84e+01 & 3.78e+01 & 0.901 & 0.903 \\
Body Volume (cm$^3$) & 1.91e+03 & \textbf{1.57e+03} & 1.74e+03 & 6.32e+02 & -1.85e+02 & 0.902 & 0.926 \\
Body $D_0$ (Gy) & 2.72e+00 & \textbf{2.50e+00} & 2.40e+00 & 1.82e-01 & 1.05e-01 & 0.899 & 0.900 \\ \bottomrule
\end{tabular}
\caption{\textbf{Results from Sparse-View CT for RT planning confirm our theoretical analysis.} For $\alpha=0.1$, 100 random calibration-test splits and symmetric (sym) and asymmetric (asym), we show a variety of RT planning metrics, calibrated mean interval lengths for biased ($L^b$) and unbiased ($L^0$) predictions, absolute bias $|b|$, mean difference between asymmetric and symmetric unbiased lengths $L_{asym}^0-L_{sym}^0$, and mean coverage ($C$).
}
\label{tab:rpa}
\end{table*}

Using this setup, we prove the following relationship for the length of a CP prediction interval using asymmetric non-conformity scores, under bias $b$:
\begin{proposition}
    \label{prop:asymIndepBias}
    Given biased predictions for a fresh test point $\hat Y_{n+1}^b = \hat Y_{n+1}^0+b$,
    the lengths for $L_1$ and CQR non-conformity scores computed using asymmetric adjustments are bias-independent: $L_{asym}(\hat Y_{n+1}^b)=L_{asym}(\hat Y_{n+1}^0)$, where $L_{asym}(\hat Y_{n+1}^b)$ and $L_{asym}(\hat Y_{n+1}^0)$ are the interval lengths computed using asymmetric adjustments for predictions with and without bias.
    
    \noindent \textbf{Proof Sketch.} By translation equivariance of $f_{lo}$ and $f_{hi}$ and the quantile operators, adding a bias $b$ to all predictions shifts both the lower and upper adjustments by exactly $b$, while the calibration scores shift in the same way. Consequently, the empirical lower and upper quantiles satisfy $q_{lo}^b = q_{lo}^0$ and $q_{hi}^b = q_{hi}^0$, and the interval span $f_{hi}(\hat Y_{n+1}^b)-f_{lo}(\hat Y_{n+1}^b) = f_{hi}(\hat Y_{n+1}^0)-f_{lo}(\hat Y_{n+1}^0)$ remains unchanged. Combining these gives $L_{asym}(\hat Y_{n+1}^b) = f_{hi}(\hat Y_{n+1}^b)-f_{lo}(\hat Y_{n+1}^b) + q_{lo}^b + q_{hi}^b= f_{hi}(\hat Y_{n+1}^0)-f_{lo}(\hat Y_{n+1}^0) + q_{lo}^0 + q_{hi}^0= L_{asym}(\hat Y_{n+1}^0)$
as claimed.
\end{proposition}

We find that asymmetric adjustments are not affected at all by a constant bias $b$, which is a desirable property. However, recall that when predictions are unbiased, asymmetric intervals tend to be longer than symmetric ones~\cite{romano2019conformalized}. 
This raises the question: what is the relationship between bias and the relative lengths of symmetric and asymmetric intervals?
We can derive a formal condition that must be true \emph{if} asymmetric intervals are observed to be shorter.
\begin{corollary}
\label{cor:SymVSAsymLengths}
For $L_1$ and CQR non-conformity scores, if the biased asymmetric interval length $L_{asym}(\hat Y_{n+1}^b)$ is shorter than or equal to the biased symmetric interval length $L_{sym}(\hat Y_{n+1}^b)$, then the following condition must hold: $2|b| \geq L_{asym}(\hat Y_{n+1}^0)-L_{sym}(\hat Y_{n+1}^0)$.
Here $b$ is the bias, and $L_{sym}(\hat Y_{n+1}^0)$ and $L_{asym}(\hat Y_{n+1}^0)$ are lengths computed using symmetric and asymmetric adjustments, respectively for hypothetical predictions without bias ($\hat Y_{n+1}^0$).

\noindent \textbf{Proof Sketch.}
We start with the premise $L_{asym}(\hat Y_{n+1}^b)\leq L_{sym}(\hat Y_{n+1}^b)$.
Substituting the equality from Prop.~\ref{prop:asymIndepBias} into the premise yields $L_{asym}(\hat Y_{n+1}^0) \leq L_{sym}(\hat Y_{n+1}^b)$.
Then, substituting the inequality from Prop.~\ref{prop:symL} gives $L_{asym}(\hat Y_{n+1}^0) \leq L_{sym}(\hat Y_{n+1}^0) + 2|b|$.
Finally, rearranging this final inequality gives the condition $2|b| \ge L_{asym}(\hat Y_{n+1}^0) - L_{sym}(\hat Y_{n+1}^0)$ as claimed.
\end{corollary}

We find that if the result ($\mathbf{R}$) $L_{asym}^b \le L_{sym}^b$ holds true, then the condition ($\mathbf{C}$) $2|b| \ge L_{asym}^0 - L_{sym}^0$ must also hold ($\mathbf{R}\Rightarrow \mathbf{C}$).
While the converse       ($\mathbf{C}\Rightarrow \mathbf{R}$) is not guaranteed, as its derivation relies on the upper-bound inequality from Prop.~\ref{prop:symL}, we later confirm that the reverse holds with high empirical validity, providing a useful heuristic for when to choose asymmetric formulations.

Inuitively, the unbiased length difference $L_{asym}^0 - L_{sym}^0$ acts as a measure of the error distribution's \emph{skew}, with a negative value indicating that the asymmetric method is already more efficient than the symmetric method. 
If asymmetric intervals are more efficient than symmetric intervals, the bias effect is large enough to overcome this initial skew difference.

\section{Experiments}
\label{sec:exps}

We evaluate Prop.~\ref{prop:symL} and \ref{prop:asymIndepBias}, Cor.~\ref{cor:SymVSAsymLengths}, and our heuristic (reverse Cor.~\ref{cor:SymVSAsymLengths}) for CQR non-conformity scores on sparse-view CT reconstruction for radiotherapy planning (RT). In scenarios with limited imaging capabilities, such as low-resource clinics~\cite{aggarwal2023radiation,court2023addressing,kisling2018radiation}, reconstruction algorithms work with observations that do not contain complete information. For example, sparse cone-beam CT algorithms use limited ($<100$ instead of the standard $100$s) 2D X-ray observations to generate 3D CT scans~\cite{sun2023ct,ying2019x2ct,shen2019patient}. The observed information is insufficient to recover the true image with complete certainty, leading to potential biases such as systematically over- or under-estimating organ volumes. 

\noindent \textbf{Setup. } We simulate a medical imaging pipeline, where a patient is imaged using sparse-CT, an image reconstruction algorithm is applied to the projections, and the resulting volume is used for downstream radiotherapy planning (RT).
We use Neural Attenuation Fields (NAF)~\cite{zha2022naf}, a self-supervised image reconstruction algorithm.
We synthetically injected noise to the projections, reconstructed the volumes using different initializations of the reconstruction algorithm, and generated plans using the Radiation Planning Assistant (RPA, FDA 510(k) cleared).

\noindent \textbf{Validation. } Following prior work~\cite{cheung2024metric}, we generate 10 reconstructions per patient for 40 patients by perturbing acquisition angles, injecting noise into the projections, and using random initializations of NAF~\cite{zha2022naf}.
We split the dataset into 35 calibration patients and 5 testing patients, and repeated the experiment 100 times with random calibration-test splits.
We use $\alpha=0.1$ for symmetric adjustments and $\alpha_{lo}=\alpha_{hi}=0.05$ for asymmetric adjustments.
We present a variety of RT planning metrics, calibrated mean interval lengths for biased ($L^b$) and unbiased ($L^0$) predictions, absolute bias $|b|$, mean difference between asymmetric and symmetric unbiased lengths $L_{asym}^0-L_{sym}^0$, and mean coverage ($C$).
The RT planning metrics are dose to 35\% relative volume of the heart (Heart $D_{35}$), heart volume, volume of right lung receiving 20Gy of dose (Right Lung $V_{20}$), dose to 35\% relative volume of the right lung (Right Lung $D_{35}$), right lung volume, left lung volume, body volume, and maximum dose to the body (Body $D_0$).
These metrics have important implications for patient safety~\cite{cheung2024metric}.

\noindent \textbf{Results. } Tab.~\ref{tab:rpa} reveals that for many downstream tasks like segmentation (Fig.~\ref{fig:seg}), predictions could be systematically biased.
This confirms that pipeline-induced bias is a measurable and highly varied problem.
In such scenarios, asymmetric formulations can improve interval length efficiency while still maintaining the target coverage.
The results also indicate that our formal theory is valid.
For all metrics, we find that Prop.~\ref{prop:symL}, Prop. ~\ref{prop:asymIndepBias}, and Cor.~\ref{cor:SymVSAsymLengths} hold true 100\% of the time on average, while the heuristic holds true 98.6\% of the time on average.

\noindent\textbf{Practical Recommendation. } Based on our results, we recommend using asymmetric formulations when residuals are biased and/or skewed. The efficiency gains over symmetric formulations are dual: they inherently capture skew, and the maximum potential improvement increases proportionally to the magnitude of the bias.

\section{Conclusion}

In this work, we argue that the effects of bias on CP prediction interval lengths with $L_1$ and CQR scores can be mitigated by computing asymmetric adjustments as opposed to the conventional symmetric adjustments.
Through experiments with RT planning, we showed the utility of asymmetric formulations when bias is present.
Ultimately, our work paves the way towards bias-aware UQ for medical imaging pipelines.


\bibliographystyle{IEEEbib}
\bibliography{refs}

\end{document}